\documentclass[sigconf,nonacm]{acmart}
\usepackage{cleveref}

\author{Claire Liang}
\email{cyl48@csail.mit.edu}
\affiliation{%
  \institution{Massachusetts Institute of Technology}
  \city{Cambridge}
  \state{Massachusetts}
  \country{USA}
}

\author{Franziska Babel}
\email{franziska.babel@liu.se}
\affiliation{%
  \institution{Link\"{o}ping University}
  \city{Link\"{o}ping}
  \country{Sweden}}

\author{Hannah Pelikan}
\email{hannah.pelikan@liu.se}
\affiliation{%
  \institution{Link\"{o}ping University}
  \city{Link\"{o}ping}
  \country{Sweden}
}

\author{Sydney Thompson}
\email{sydney.thompson@yale.edu}
\affiliation{%
 \institution{Yale University}
 \city{New Haven}
 \state{Connecticut}
 \country{USA}}

\author{Xiang Zhi Tan}
\email{zhi.tan@northeastern.edu}
\affiliation{%
  \institution{Northeastern University}
  \city{Boston}
  \state{Massachusetts}
  \country{USA}
}

\usepackage{cuted}
\usepackage{capt-of}
\usepackage{subcaption}
\usepackage{fancyvrb}
\usepackage{gensymb}
\usepackage{float}
\usepackage{stfloats}
\usepackage{tikz}
\usepackage{xcolor}

\usepackage{soul}

\definecolor{yellow}{RGB}{255, 255, 209}
\definecolor{orange}{RGB}{255, 204, 204}
\DeclareRobustCommand{\hlpink}[1]{{\sethlcolor{pink}\hl{#1}}}
\DeclareRobustCommand{\hlx}[1]{{\sethlcolor{yellow}\hl{#1}}}

\begin{document}

\title{A Checklist for Deploying Robots in Public: Articulating Tacit Knowledge in the HRI Community}

\renewcommand{\shortauthors}{}

\begin{abstract}
Many of the challenges encountered in in-the-wild public deployments of robots remain undocumented despite sharing many common pitfalls. This creates a high barrier of entry and results in repetition of avoidable mistakes. To articulate the tacit knowledge in the HRI community, this paper presents a guideline in the form of a \textit{checklist} to support researchers in preparing for robot deployments in public. Drawing on their own experience with public robot deployments, the research team collected essential topics to consider in public HRI research. These topics are represented as modular flip cards in a hierarchical table, structured into deployment phases and important domains. We interviewed six interdisciplinary researchers with expertise in public HRI and show how including community input refines the checklist. We further show the checklist in action in context of real public studies. Finally, we contribute the checklist as an open-source, customizable community resource that both collects joint expertise for continual evolution and is usable as a list, set of cards, and an interactive web tool.

\end{abstract}

\ccsdesc[500]{Computer systems organization~Robotics}
\ccsdesc[500]{Human-centered computing}
\ccsdesc[500]{Human-centered computing~Human computer interaction (HCI)}

\keywords{public robots, system deployment, tools}

\begin{teaserfigure}
    \centering
    \includegraphics[width=0.9\textwidth]{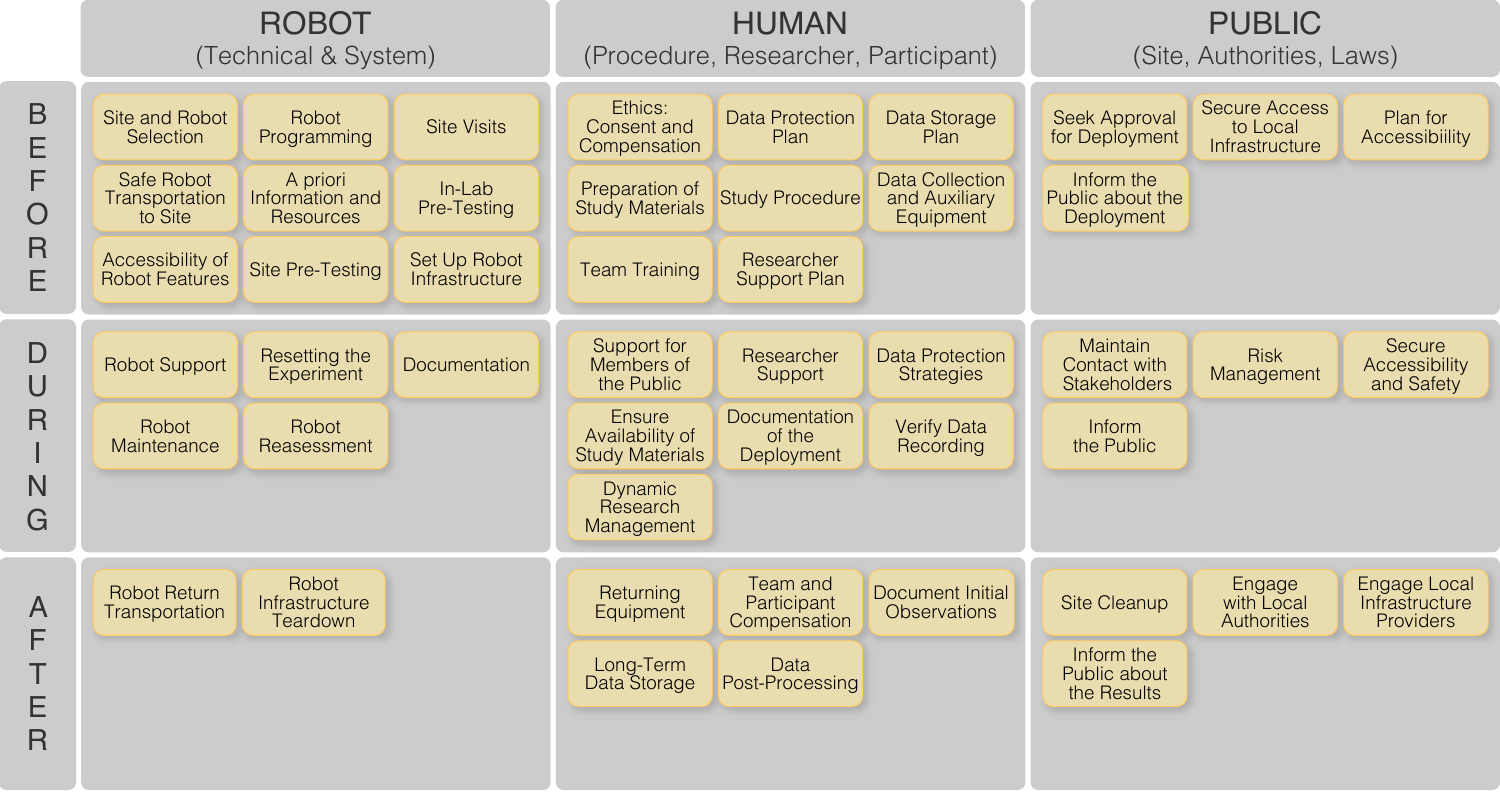}
    \caption{A checklist for robot deployments in public. Temporal phases (\texttt{Before}, \texttt{During}, \texttt{After}) are represented as table rows and domains (\texttt{Robot}, \texttt{Human}, \texttt{Public}) are represented as table columns. Topics to consider for each phase-domain combination are represented by yellow flip cards.}
    \Description{All cards are described in the text in~\cref{sec:checklist-explanation}. The card topics in each cell read as follows (listed from top to bottom,  left to right): Column Robot contains the following cards. Before: Site and Robot Selection, Robot Programming, Site Visits, Safe Robot Transportation to Site, A priori Information and Resources, In-Lab Pre-Testing, Accessibility of Robot Features, Site Pre-Testing, Set Up Robot Infrastructure. During: Robot Support, Resetting the Experiment, Documentation, Robot Maintenance, Robot Reassessment. After: Robot Return Transportation, Robot Infrastructure Teardown. Column Human contains the following cards. Before: Ethics: Consent and Compensation, Data Protection Plan, Data Storage Plan, Preparation of Study Materials, Study Procedure, Data Collection and Auxiliary Equipment, Team Training, Researcher Support Plan. During: Support for Members of the Public, Researcher Support, Data Protection Strategies, Ensure Availability of Study Materials, Documentation of the Deployment, Verify Data Recording, Dynamic Research Management. After: Returning Equipment, Team and Participant Compensation, Document Initial Observations, Long-Term Data Storage, Data Post-Processing. Column Public contains the following cards. Before: Seek Approval for Deployment, Secure Access to Local Infrastructure, Plan for Accessibility, Inform the Public about the Deployment. During: Maintain Contact with Stakeholders, Risk Management, Secure Accessibility and Safety, Inform the Public. After: Site Cleanup, Engage with Local Authorities, Engage Local Infrastructure Providers, Inform the Public about the Results.}
    \label{fig:checklist}
\end{teaserfigure}

\maketitle

\section{Introduction}
\label{sec:intro}
To fully integrate robots into our everyday lives, real-world deployments, although challenging, are necessary in HRI research. 
Conducting studies in real spaces allows us to demonstrate the efficacy and benefits of our robot systems in the real world. In particular, HRI researchers that conduct those studies are often interested in the natural interactions that can only arise when passersby encounter a robot in the wild \cite{Pelikan2024,Babel2022, Brsci2015}.

Public and semi-public spaces (hereafter referred to as "public") are dynamic environments characterized by broad access, making them largely outside the researcher's control. The resulting difficulty is exacerbated when every site is unique, encounters are largely unrestricted, and there is no standardized practice for approaching deployments. 
Real world deployments are challenging, even for experienced researchers, will never be flawless and can never be fully planned for.
When preparing for and executing a public deployment, researchers need to juggle logistics, technical development, unique ethical considerations, and a multitude of unexpected events-- perhaps even antagonistic behaviors towards the robot~\cite{Sabanovic2006,Babel2022,Yamada2020}.
Finding opportunities to share our experiences has become increasingly valuable to the community in recent years, as demonstrated with the organization of workshops such as ``Real-World HRI in Public and Private Spaces: Successes, Failures, and Lessons Learned'' \cite{pubrob2025} and``Human-Robot Interaction in Extreme and Challenging Environments'' \cite{wild2025}. These workshops created spaces where researchers could share anecdotes and insights. 

However, outside of sharing anecdotes in conversations with other researchers, there is currently no unified resource for HRI researchers to share their experiences and lessons learned to the broader community. The main means of research dissemination--- publication---does not incentivize submissions that focus on negative results or lessons learned. As a result, many research groups unnecessarily repeat one another's mistakes due to the lack of HRI community-wide communication. %
This creates an avoidable high barrier to entry for public robotics research, which only further exacerbates the lack of shared knowledge. This especially noticeable for young HRI researchers and those new to the field, as the current approach is to rely on institutional knowledge. %
Part of what makes unifying HRI research in public so challenging is its unique relationship with the ``unexpected'' - no amount of preparation can prepare a researcher for every scenario they will encounter.
Furthermore, researchers in public HRI represent an enormous variety of research objective, priorities, and stakeholders.

To address this gap, we developed a guideline for robot deployments in public in the form of a checklist. This checklist is rooted in experiences from researchers in the public HRI community, seeded \footnote{This paper's checklist version has been developed  with these six authors in the window of the KTH RPL Summer School in 2022 through September 30th, 2025}{initially by the six interdisciplinary HRI authors on this paper} and further nuanced through interviews with six researchers of different career stages, educational backgrounds and deployment experiences in a range of public settings. While we build on \textit{checklist items} like a traditional checklist, we instill a hierarchical structure that permits flexibility and personalization. We group checklist items that are based on real deployment experiences into \textit{topics}, which are respectively on the back and front of \textit{flip cards}. These topics, in the form of flip cards, are sorted into a \textit{table} structure where they are organized by the temporal phase (before, during, after) in which they occur during a deployment, and what domain (robot, human, public) they are relevant to. \Cref{fig:checklist} illustrates the checklist in this format.

We demonstrate how the checklist can be used by HRI researchers and discuss how it can be further expanded. The topic flip card construction that unites individual checklist items makes the resource accessible for long term community-wide engagement since checklist items can be easily added to the back of a card. We host our resource on a website\footnote{\url{https://hripublicdeploymentchecklist.github.io/}} where the checklist can be customized and used directly or downloaded as a Microsoft Word file.

The contribution of our paper is threefold: First, we contribute a guideline for public robot deployments, grouping checklist items into  topics that are represented by flip cards with an imposed organization into deployment phase and domain. Second, we provide a customizable community resource that researchers can review, personalize and use as a checklist for their next robot deployment in public. Third, we demonstrate how the checklist can be used in action, demonstrating our vision for the long-term application of this resource and potential for community-steered evolution.

\section{Related Work}
\label{sec:relwork}
We introduce work on robot deployments in public and provide an overview of tools for articulating knowledge in the HRI community.
\subsection{Public Robot Deployments in HRI}
\label{sec:relwork-pubrobdeployments}
To capture natural behaviors with the general public, many researchers embrace the practice of performing in-the-wild research for system evaluation with ecological validity~\cite{Liang2023}, testing methods of robot behavior generation~\cite{mavrogiannis2023core,du2024can}, identifying interesting human behaviors~\cite{taylor2022regroup,thompson2024predicting} or pursuing community centered design~\cite{winkle2021leador}.

Previous deployments have targeted a broad range of public spaces (museums~\cite{Sabanovic2006, Shiomi2006, Pitsch2009}, malls~\cite{Brsci2015, Kanda2010}, and transportation hubs~\cite{Babel2022, thunberg2020, joosse2017}) and semi-public environments (universities~\cite{gockley2005designing, Thompson2024, Liang2023, Tan2024}). 
They have even expanded into outdoor spaces like city centers~\cite{Bu2024, Brown2024} and sidewalks~\cite{Pelikan2024, Dobrosovestnovaa, reig2018field, Weinberg2023} where researchers investigated the effect of robots (e.g., trash-collecting \cite{Bu2024,Brown2024} or delivery robots \cite{Abrams2021}) on the everyday lives of laypeople.
\par
Robot deployments in public spaces face unstructured, dynamic, and crowded environments with interactions ranging from typical to unusual~\cite{Pelikan2025, Tan2024, Babel2022}.
Some of these unusual events are largely unique to public spaces, such as very brief and singular encounters as a symptom of members of the public simply sharing the same space with the robot without even intending to interact with them, a population that is commonplace in public deployments~\cite{Dobrosovestnova2025}.
Brief encounters of this variety may lead to unexpected behaviors. For instance, service robots in public might be perceived as unpredictable due to robot novelty or lack of communication behaviors~\cite{Chadalavada2015, Matthews2020}. %
These unusual experiences can be observed through phenomena like sudden, abrupt stops from the robot (`the halting problem' \cite{brown2023}) causing startled, evasive passerby actions \cite{Babel2022}.

For experienced public HRI researchers, these are known problems. With experience, after personally overcoming these unexpected behaviors, failures, and errors, they become part of the expert's tacit knowledge. Yet, because it is not shared with other researchers in the community through conferences and journals, that wisdom remains silent. We believe that those lessons-learned are highly valuable and work towards converting this knowledge to create a concrete resource for the HRI community. 

\subsection{Existing Guidelines and Recommendations}
\label{sec:relwork-existing}

The HRI research community has long worked towards identifying best practices by providing community resources and guidance. An early example is Steinfeld et. al.'s common metrics paper~\cite{steinfeld2006common}, which identified metrics that are useful for HRI studies. 

A recent multi-lab initiative \cite{saad2025tutorial} has created an online resource for collecting and recommending common HRI questionnaires. Beyond metrics, the community has also embraced guidelines for conducting research and reporting. Hoffman and Zhao wrote a primer on how to conduct empirical research in Human-Robot Interaction~\cite{hoffman2020primer}. Riek surveyed HRI papers on their usage of wizard-of-oz and recommended guidelines on how such studies should be reported to increase rigor and reproducibility~\cite{riek2012wizard}. Winkle and colleagues elucidate a method for conducting end-to-end participatory design in HRI \cite{winkle2021leador}. Junior researchers can look to this body of work for guidance on many aspects of HRI work.
\par
The subcommunity of HRI researchers who conduct studies in public spaces are currently benefiting from new theories and methods. These include theories about what forms of interaction are being investigated in public spaces (i.e. "Existence Acceptance") \cite{Abrams2021}, how to categorize the people interacting with the robot \cite{Dobrosovestnova2025, Rosenthal-VonDerPutten2020} and how to approach design of robots for public spaces, grounded in ethnographic observations \cite{Pelikan2025}. Furthermore, Kraus and colleagues present a checklist and recommendations for acceptable and trustworthy design of HRI which also contains aspects for public deployments~\cite{kraus2022trustworthy}. Salvini and colleagues list design recommendations to ensure continued safe operation during public robot deployments (e.g., expecting vandalism, bullying) \cite{Salvini2010}.
\par
However, to the knowledge of the authors, we do not yet have resources on the practicalities of deploying a robot in public. The actual day-to-day process goes relatively unguided, especially if the research team does not include an experienced public HRI researcher. Our checklist centers on the hands-on parts of public deployments-- what steps need to be taken-- to address the details that collectively that comprise some of the most tedious and challenging parts of putting a robot in a real world public space.

\section{Method}
\label{sec:method}
The checklist we present in this paper is the result of discussions by the author team about their own public deployment experiences. After sharing common pitfalls and unexpected events, we moved towards a systematic collection of these into a checklist that combines our experience and know-how. We iterated on the checklist and eventually converged to a modular flip card format. We then interviewed six experts to gather their experiences and their commentary on the checklist, integrating their feedback into the checklist to further transform it into an even more expansive iteration.

\subsection{The Authors' Deployment Experience}
\label{sec:method-authorexperience}

We, the author team, unite backgrounds that range from computer science and robotics to psychology and interaction studies. While there are considerable differences in our backgrounds and research philosophies, we all share experience and interest in bringing HRI work into public spaces. Collectively, we have 42 years of experience working on deploying robots in the wild across North America and Europe.
In discussing and contrasting our experiences, we discovered common patterns, saw the potential value to the community, and explored how to document and share our accumulated knowledge.
Critically, we realized that while some failures were inevitable, in many cases if we had known to prepare for specific situations based on the knowledge of other team members, we would have experienced far fewer failures, delays, and setbacks during the deployment. 

In 2024, we decided to compare and contrast our insights from our own projects and distill our shared experiences into something we would want for our past-selves when we were first embarking on HRI in public. Since the root cause of a failure often originated in the preparation of the project, we also expanded our lens beyond during the phase of an ongoing deployment. We mapped our entire process of planning and carrying out robot deployments in public, envisioning a meta-checklist to help other researchers implement their own deployment plans. Our efforts can be seen as a way to articulate knowledge in the community into a tangible format that others can draw upon and adapt for their own purposes \cite{Lupettietal2021,Pelikan2025}.

\subsection{Creating the Initial Checklist}
\label{sec:method-creating}
We developed the checklist in three iterations -- we started with a compiled list, followed by a table form, and finally a table of flip cards.

\subsubsection{List}
\label{sec:method-creating-list}
First, all authors reflected on their own robot deployments and created lists of steps they took (or wish they had taken) in conducting their deployments. These steps included experiences of failures that were relevant and how they could have been avoided. Subsequently, we collaboratively combined our lists, collapsing similar bullet points and discussing aspects that appeared unique to specific deployments. The list loosely followed three different \textbf{phases}, sorting aspects to consider \texttt{Before}, \texttt{During} and \texttt{After} a robot deployment in public. 

\subsubsection{Table} 
\label{sec:method-creating-table}
Collecting all our experiences culminated in an extensive, hard to read list. To find a way to bring structure into it, we grouped the items into three different \textbf{domains} loosely sorted into \texttt{Robot} related aspects that are mostly of a technical and systems nature, aspects related to the \texttt{Human}, primarily building on a user studies perspective and aspects that are related to the specific site, and working in a \texttt{Public} setting where researchers may need to consider other stakeholders. 

Building on the prior step, we organized our checklist into a 3x3 table format, with \textbf{phases} as rows and \textbf{domains} as columns. The table presentation not only made the information more easily digestible but also allowed the members of the author team to focus on areas where they had more relevant experience.

\subsubsection{Flip Cards} 
\label{sec:method-creating-cards}
As the checklist grew, it quickly became overwhelming and unnavigable. We discovered that we had to make further divisions, introducing subheadings within each cell of the table. Searching for and identifying relevant steps to prepare for a deployment was difficult in such a format. 

\begin{figure}[h]
    \centering
    \includegraphics[width=.9\linewidth]{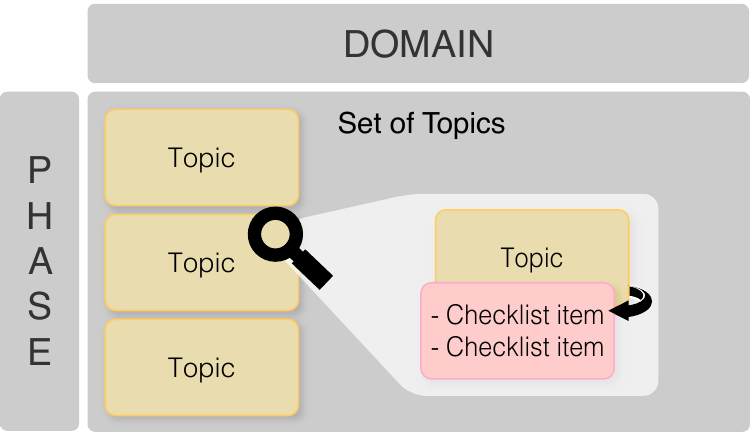}
    \caption{The hierarchical structure of the checklist. For each combination of domain (column) and phase (row), a set of topics is represented as flip cards. Each flip card has the topic name on the front side and the checklist items on the back.}
    \label{fig:hierarchy}
\end{figure}

Eventually, we had the idea to structure our checklist into high-level \textbf{topics} and low-level \textbf{checklist items}. Exploring how to hide low-level details but making them available if necessary, we arrived at the format of a \textit{flip card}, with the front showing the topic name and the back listing more detailed items of the checklist. The structure is illustrated in \autoref{fig:hierarchy}.

This format allowed us to collect \textbf{sets of topics} in each table cell. This results in an overview of the 48 card topics collected in total, while each of them can be flipped individually to reveal further details about the items that fall under this topic. We found the metaphor of flip cards particularly suitable since it supports our vision of researchers assembling a personalized checklist from our card deck. One could review the entire card deck first, and then pick cards from the deck that are most relevant to them. 

\subsubsection{Implementation} 
\label{sec:method-creating-implementation}
We implemented the first version of our checklist through a Miro board\footnote{https://miro.com} using their flip card feature. The flip card format enabled us to see the whole table while working on individual cards by zooming in and out. Over the summer of 2025, we performed a series of review sessions as a team to reach consensus on which cards to use, what level of specificity should be covered, what to merge, and what to differentiate. %

For easy access by the HRI community, we translated our flip card table from Miro into a website. The website has two parts -- the \textbf{flip cards} displaying \textbf{topics} arranged in the table on the left, and an information sheet on the right. When a visitor clicks on a card, the card is flipped, and the \textbf{checklist items} along with a description of the topic are displayed in the information sheet. 

There are two ways for any member of the community to contribute to the checklist. First, our website will be made open source and shared on GitHub. Researchers can branch from our repository and personalize the checklist for their specific use case and share with others if they so choose. Second, if a researcher wants to make suggestions to modify and improve the checklist as a whole, they can use github's functionality to ``raise an issue'' which essentially leaves a comment including their suggestions. This allows the community to see each other's suggestions asynchronously without being restricted to workshops and in-person events.

For ease of use, the website also includes functionality for visitors to personalize their checklist and download it as a Microsoft Word document to use for a deployment.

\subsection{Expert Interviews}
\label{sec:method-expertinterviews}
After creating our initial checklist, we conducted a set of semi-structured expert interviews to understand how this checklist would serve the real public HRI community. Each interview consisted of two phases: gathering context about the expert's research experience and collaboratively exploring the checklist.
We were interested in learning how the checklist could be used as a community tool, if the checklist structure and format were intuitive, and whether some items were missed or appeared misplaced. %

In the first portion of the interview, we asked the experts to provide an example of one of their prior robot deployments in public giving us a vivid picture about the location of the study, the research team, and the duration of the deployment. We also asked whether they used any of their own checklists during the study, and if there were any things they would do differently. For the second portion of the interview, experts shared their screen while exploring our checklist in a tabular form similar to Figure \ref{fig:checklist}, allowing us to observe how they interacted with it. We briefly explained the structure of the checklist and gave the experts the option to select a specific subsection to examine in more detail, or to review the checklist more holistically. As they explored, we asked them to point out  aspects that were new to them, elements that may be missing, and to share experiences that resonated with the checklist topics. Beyond helping us to further refine the checklist, we wanted to explore a potential future use of the checklist, where the community can continue to engage with the resource and further evolve it.

\begin{figure*}[ht!]
    \centering
    \includegraphics[width=\linewidth]{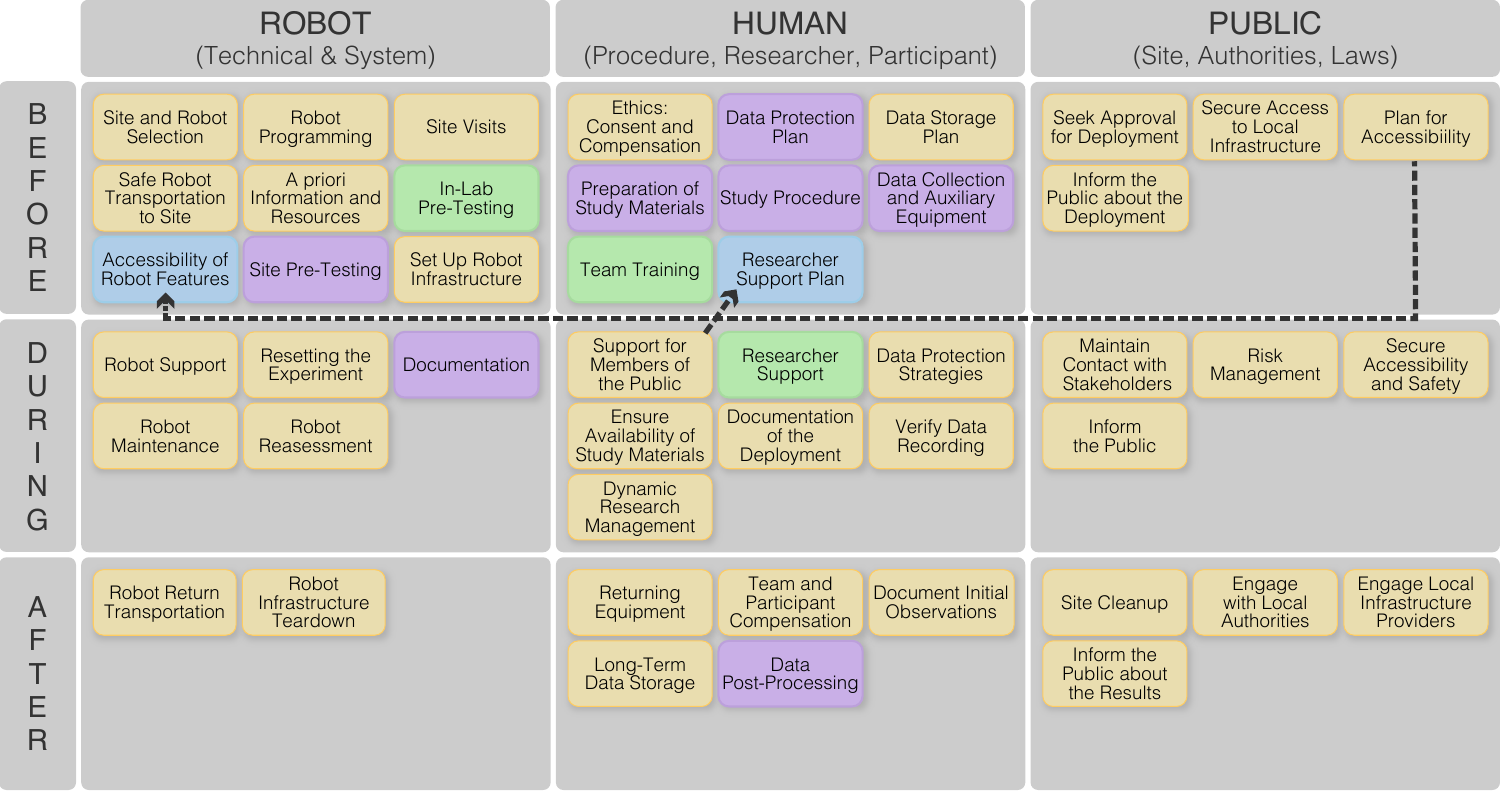}
    \caption{A depiction of changes done to the checklist after the interviews. Green: card added based on feedback from P1 and P3, including checklist items on back; Purple: items were made by P2; Blue: new card formed based on responses to existing card from P2 and P3. The original card responded to is indicated with arrow.}
    \label{fig:changes}
\end{figure*}

We recruited six experts keeping diversity of backgrounds, geographic location, and level of seniority in mind. The interviews lasted 60 minutes and were conducted via Zoom. We obtained the necessary ethical approvals before the interviews. The interview data is stored on servers in the approved venues. All six expert interviews were reviewed by the author team and we used their feedback to update the checklist. %
Most changes or suggestions were about adding items to the checklist. Some topics or checklist items required further clarification. For those topics and items, we altered wording to better convey the intended meaning, and in some cases added additional checklist items to a topic. \autoref{fig:changes} visually depicts all of the changes we made during the process as a summary. %

\section{The Checklist}
\label{sec:thechecklist}
The final checklist, presented in Figure \ref{fig:checklist}, is structured as a 3x3 table. Each \textbf{domain} - the \texttt{Robot}, the \texttt{Human}, and the \texttt{Public} - is represented by a column. Each \textbf{phase} -  \texttt{Before}, \texttt{During}, and \texttt{After} - is represented by a row. As shown in Figure \ref{fig:hierarchy}, each combination of domain and phase contains a set of topics, each represented by a flip card. The \textbf{topic} name (labeled in \hlx{\texttt{yellow}}) is listed on the front of the flip card and the \textbf{checklist items} for that topic are on the back of the card (labeled in \hlpink{\textit{pink}}).

To guide researchers in using the checklist all domain-phase combinations are described in~\Cref{sec:thechecklist-explanation} and a selection of topics and their checklist items are described in Section~\cref{sec:thechecklist-flippingcards}. Our aim is for the set of all topics to cover a large part of how to prepare for a deployment, but not every topic or checklist item may be necessary for every deployment. We provide an examples on how the checklist can be used and customized in Section~\cref{sec:thechecklist-inaction}, building on the interviews and on Sydney's own experience.

\subsection{The Checklist Topics}
\label{sec:thechecklist-explanation}
We will present the checklist by domains (from left to right in~\cref{fig:checklist}), for ease of understanding. We suggest researchers should review every topic and consult the specific checklist items on the back of the cards as relevant for their own purposes. 

\subsubsection{Domain: Robot}
\label{sec:thechecklist-explanation-robot}

The set of topics in \textbf{\texttt{Robot-Before}} concerns how to prepare the robot to operate in a relatively unstructured environment. Safety is of utmost priority, and iterative testing both in-lab and on-site is necessary to have sufficient time to adjust to emerging problems and to make relevant changes.

Often, the first step is \hlx{\texttt{Site and Robot Selection}}. The choice of robot and site are heavily dependent on one another. For example, a robot with small wheels would be a poor choice for a cobbled public square, but may perform well in a museum with smooth marble floors.
In terms of basic preparations,
\hlx{\texttt{Robot Programming}} will need to be done with particular attention to moving safely. One should perform in-person \hlx{\texttt{Site Visits}} to ensure  feasibility of the deployment. When selecting the appropriate robot and site, the research team also must ensure \hlx{\texttt{Safe Robot Transportation to the Site}}, making sure all equipment can be moved there. During site visits, the researcher must gather the necessary \hlx{\texttt{A priori Information and Resources}} to ensure robot functionality---for example, this may include mapping the space for a navigation robot or preparing software to monitor the robot. Researchers should engage in as much \hlx{\texttt{In-Lab Pre-Testing}} as possible before bringing the robot to the deployment site. This may include simulation. By integrating knowledge about the site and robot during development, researchers may be able work for better \hlx{\texttt{Accessibility of Robot Features}}, for instance by considering adjustable volume and clear visuals that also work in bright light. Once the robot is sufficiently safe and ready for transportation, \hlx{\texttt{Site Pre-Testing}} with the robot is absolutely crucial to prepare for contingencies that can emerge outside the lab and to ensure a successful deployment. Based on the observations made during the pre-test, researchers can also  \hlx{\texttt{Set Up Robot Infrastructure}}, including finding suitable charging locations. 

The set of topics in \textbf{\texttt{Robot–During}} 
describes the monitoring work for consistent performance of the many components in the robot system to ensure expected behavior. Topics include the scheduling and planning of \hlx{\texttt{Robot Support}}, who is physically present to handle the robot and what they should do when errors occur. Procedures for \hlx{\texttt{Resetting the Experiment}} between sessions can be helpful. Any changes or corrections should be covered in \hlx{\texttt{Documentation}} to ensure continuous performance. Regular \hlx{\texttt{Robot Maintenance}} of all parts of the system to ensure it runs safely and \hlx{\texttt{Robot Reassessment}} to identify potential errors is crucial.

The set of topics in \textbf{\texttt{Robot-After}} is smaller but just as relevant, as it covers concerns at the end of the deployment when returning back to the lab. The researcher will be in charge of removing the robot from the space, ensuring \hlx{\texttt{Robot Return Transportation}} in a safe and orderly manner. In addition, the \hlx{\texttt{Robot Infrastructure Teardown}} involves removing all equipment related to the robot and returning it to its storage location, checking that the robot is in good condition, and the data it collected is handled appropriately. %

Considering these topics allows technical researchers to manage the robot system in context of the site parameters when the deployment time comes and provides interdisciplinary research teams with an overview of the preparatory work before a deployment.

\subsubsection{Domain: Human}
\label{sec:thechecklist-explanation-human}

The set of topics in \textbf{\texttt{Human-Before}} covers aspects that need to be considered before any human physically encounters the robot in the context of the research study. Researchers must prepare to keep people safe and protected, including both the humans encountering the robot at the deployment site and research team members. 

Like in any user study, \hlx{\texttt{Ethics: Consent and Compensation}} must be planned early in the deployment preparation. In addition, having a \hlx{\texttt{Data Protection Plan}} and a \hlx{\texttt{Data Storage Plan}} in advance is helpful in handling the specific challenges that emerge from studying and recording people in public. Researchers should consider the \hlx{\texttt{Preparation of Study Materials}} such as scripts and protocols. Planning the \hlx{\texttt{Study Procedure}} can help to achieve a smooth deployment and to support research team members. Such plans should be carefully reviewed by the researchers as well as any ethics review processes, such as a university's internal review board, before pilot testing at the real site begins. The research team should plan for and practice with \hlx{\texttt{Data Collection and Auxilliary Equipment}}, including the use of cameras and tripods to document the interactions that people have with the robot. Researchers must receive sufficient \hlx{\texttt{Team Training}} for running the study. They should also be primed for the personal impact fieldwork may cause, which can be considered in a \hlx{\texttt{Researcher Support Plan}}–––as real people bring real emotions, researchers may want to prepare themselves for getting impacted by the deployment in unexpected ways. 

The set of topics in \textbf{\texttt{Human-During}} covers aspects related to the ongoing research study and the people in the space. This includes continued \hlx{\texttt{Support for Members of the Public}} and \hlx{\texttt{Researcher Support}} following the respective plans and making adjustments as necessary. Signs informing the public about \hlx{\texttt{Data Protection Strategies}} should be continuously evaluated and moved if they are not sufficiently visible. Researchers need to \hlx{\texttt{Ensure Availability of Study Materials}} at all times, which sometimes just means printing extra forms and sometimes demands something more complex. Furthermore, researchers should consider the \hlx{\texttt{Documentation of the Deployment}}, which can include video recording or making notes on any changes to the study as well as unexpected or unusual human behaviors, as these can lead to interesting new discoveries. Much like robot system management, researchers should continuously \hlx{\texttt{Verify Data Recording}}, making sure that any equipment is running as planned---it is not uncommon to lose a session or two from a study because someone forgot start a recording or a battery ran out. Researchers should also be generally open to adjustments to procedures as needed, engaging in \hlx{\texttt{Dynamic Research Management}} such as adapting locations for on-site interviews when noticing that they are interfering with the natural dynamics of the site. 

Finally, the set of topics in \textbf{\texttt{Human-After}} includes aspects that are related to withdrawing from the study site. This includes removing and \hlx{\texttt{Returning Equipment}} that has been used to document people's interactions with the robot such as cameras. After a deployment is complete, researchers should consider \hlx{\texttt{Team and Participant Compensation}} in a timely fashion if people have provided help. Depending on the specific deployment, this can be monetary compensation or a smaller form of recognition such as having an informal coffee gathering with the whole team. Such opportunities can also be used to \hlx{\texttt{Document Initial Observations}}, jotting down interesting observations while the experience is still fresh. Right after the study is also the optimal time to set up \hlx{\texttt{Long-Term Data Storage}} by transferring and collecting files for later analysis as well as starting \hlx{\texttt{Data Post-Processing}}, for instance blurring faces from videos before off-board file transfer. 

\subsubsection{Domain: Public}
\label{sec:thechecklist-explanation-public}

The set of topics in \textbf{\texttt{Public-Before}} centers around acknowledging that the research team and robot system are guests in other people's space. Researchers should build strong relationships with relevant third parties -- positive relationships can help the deployment go much more smoothly, especially if something goes wrong or the project timeline needs to be adjusted. This can mean that researchers need to \hlx{\texttt{Seek Approval for Deployment from Site Authorities}} and \hlx{\texttt{Secure Access to Local Infrastructure}}, such as internet connectivity. In exchange, researchers must be respectful of the space and \hlx{\texttt{Plan for Accessibility}}, for instance by ensuring that emergency exits are not obstructed. A good practice is to \hlx{\texttt{Inform the Public About the Deployment}}, whether to manage expectations or to recruit more people to engage with the robot.

The set of topics in \textbf{\texttt{Public-During}} are concerned with continued engagement with the people present in the public space during the deployment. The third parties that the researchers have established a rapport with before should be maintained if not strengthened throughout the deployment. This can mean researchers should \hlx{\texttt{Maintain Contact with Stakeholders}} while being present in the site, continuing an ongoing dialogue and further engaging in \hlx{\texttt{Risk Management}} as concerns might come up during the deployment. Being in close contact with janitors, security guards or other people working in public can be especially relevant. Frequent reassessment of the plans to \hlx{\texttt{Secure Accessibility and Safety}} is crucial, as it is typical in public spaces for obstacles to move and conditions to change which requires modification to the robot's behavior. Researchers also must be ready to continuously \hlx{\texttt{Inform the Public}} during the deployment. Researchers will often need to answer questions about the robot and the study as people pass through the public site. 

Finally, the set of topics in \textbf{\texttt{Public-After}} concerns aspects to consider after withdrawing from the public site. After the deployment is complete and the \hlx{\texttt{Site Cleanup}} is finished, researchers should continue to \hlx{\texttt{Engage with Local Authorities}}, following up on results and answering emerging questions. Similarly, it is valuable to \hlx{\texttt{Engage Local Infrastructure Providers}} as they play a huge role in successful deployments. %
Following up and finding ways to \hlx{\texttt{Inform the Public about the Results}} can also be a valuable way to let the people who were studied in public to know about the insights gained from the research.

\subsection{Flipping the Cards: Checklist Items}
\label{sec:thechecklist-flippingcards}
Each topic corresponds to a set of checklist items, which are represented on the back of the flip card, as shown in Figure \ref{fig:hierarchy}. In this subsection, we hone in on three example topics, and contextualize the checklist items that are provided on the back of those flip cards. A full list of topics, checklist items and explanations like the ones included here can be found %
by using our web tool\footnote{\url{https://hripublicdeploymentchecklist.github.io/}}.

\subsubsection{Site Visits}
In the \texttt{Robot-Before} set, we take a closer look at the topic \hlx{\texttt{Site Visits}}. The checklist items on the back of the card cover the following aspects.

Before the robot is brought to the site, research team members should visit the site to \hlpink{\textit{Assess whether the robot's task and research question can reasonably be accomplished at the site}}, i.e. whether it is suitable for their project. Aspects to consider are whether the research questions, the robot's task, and/or the robot are suitable for the site. For instance, a robot that relies on hearing spoken responses from members of the public cannot be deployed at a noisy festival. Ideally, researchers should \hlpink{\textit{Visit site at similar time to expected deployment}}. For instance, visiting a transit station at night when the study is intended to occur during the day will provide an unrepresentative sense of what challenges the researchers should anticipate. If researchers \hlpink{\textit{Explore how the robot will move in the space}}, they will better understand how the robot will interface with the physical environment around it.  Doing such tests in the site allows researchers to \hlpink{\textit{Identify potentially challenging areas for locomotion or sensing}} (tactile paving, gutters, staircases) that a robot may not be able to navigate over, or for a stationary robot, a location where the robot does not have sufficient lighting for its sensors. Common examples worthy of their own checklist items are for researchers to \hlpink{\textit{Try out Wi-Fi connection/setup in multiple locations}} as well as to \hlpink{\textit{Locate power sources for charging during the day}} while the robot is in the deployment site to avoid running out of battery.

\subsubsection{Support for Members of the Public}
\label{thechecklist-flippingcards-participantsupport}
In the \texttt{Human-During} set, we take a closer look at the topic \hlx{\texttt{Support for Members of the Public}}. The checklist items on the back of the card consist of two main components.

When conducting a public HRI study, the researcher needs to anticipate and adapt to respond to people in public settings. The researchers should \hlpink{\textit{Prepare ways for members of the public to contact researchers}} in case members of the public would like to make requests to the researchers directly, such as having their data removed after their robot encounter. It is also valuable to \hlpink{\textit{Prepare for frequently asked questions}} so that robot handlers are ready to deal with emerging questions. A typical example would be having an answer ready to the question, ``What is the robot doing?'', explaining the purpose of the study and the functionalities of the robot. As the study progresses, more frequently asked questions can be added to the list. To ensure that the communication between the site authorities, robot handlers and members of the public is correct, consistent, and will not create confounds for the study, it is beneficial to write down brief statements beforehand. 

\subsubsection{Engage with Local Authorities}
\label{thechecklist-flippingcards-engage}
In the \texttt{Public-After} set, under the topic \hlx{\texttt{Engage with Local Authorities}} we examine two checklist items.

Part of conducting a public HRI study will always be the camaraderie or rapport a researcher may establish with people such as building managers or local IT professionals. It is good practice to express gratitude, and researchers may want to \hlpink{\textit{Write thank you notes/emails and seek support of future collaboration to with local authorities}} after the end of the deployment. For some researchers who hope to continue engaging with the same communities it can be important to continue to foster and support these relationships. It is also typical for researchers to \hlpink{\textit{Send results if desired}}. In some cases, this may be explicitly requested by stakeholders (e.g. an on-site staff member who would like to see final publications or outcomes). Depending on the site this may also include presenting intermediate results or sharing the end product before it is published.

\subsection{The Checklist in Action}
\label{sec:thechecklist-inaction}

We envision the checklist being used in three primary ways: First, individual researchers can use the checklist at the \hlx{\textit{topic}} level of granularity as a guideline for planning a public robot study.  Second, by personalizing the checklist \hlpink{\textit{items}}, a researcher can export concrete list of steps to follow when preparing and executing their specific robot deployment. Third, members of the community can improve and refine the general checklist structure for other community members' use.

\subsubsection{Topics Guiding Deployment in Action}
\label{sec:thechecklist-inaction-topics}

\footnote{The study was conducted in 2025}{Sydney conducted a robot deployment in a publicly accessible rotunda at a university after the checklist was completed}. They had deployed the same robot multiple times previously, but entered a new site and were working with new robot handlers. This presented fresh challenges, and a good opportunity to put the checklist to the test. Sydney referred to the checklist at the topic level as a guideline for planning and preparation and noted their experience throughout the process. 

The topic \hlx{\texttt{Site Administration}} prompted Sydney in the process of coordinating with building administration. In their study, event coordinators required much more granular information about the project plan than in Sydney's prior experiences. By referencing the checklist topic, Sydney was able anticipate the concerns of the the building administration, by using the accumulated experience from other researchers via the checklist before encountering any pushback thus expediting the approval process. %

Furthermore, the topic \hlx{\texttt{Study Procedure}} could inspire Sydney to create ``how to start the robot'' and ``what to do when things go wrong'' guides that the robot handlers used to perform simple troubleshooting on site without Sydney's direct supervision. The \hlx{\texttt{Documentation of the Deployment}} topic also served as a reminder for robot handlers to document interesting phenomena during the deployment. One such observation was that on a rainy day, more people than usual were in the rotunda. People spent more time standing inside to wait out the rain instead of passing through as normal. A small change in weather represented a major change in human behavior, which made this phenomenon and its cause important context to note for future use of the dataset. Without the checklist reminder to teach the handlers how to recognize relevant changes in behavior patterns, this might not have been noted.

\subsubsection{Checklist Item Personalization in Action}
\label{sec:thechecklist-inaction-items}
The checklist is intended as a community resource and we envision it to continue to evolve as the HRI community further engages with it. All six public HRI researchers we consulted considered the checklist to be compelling and potentially useful for themselves and for the wider community. They generally understood the checklist format and could navigate it without further instructions.

During our conversation with P2, they reviewed each flip cards individually and organically personalized the checklist items without our prompting. P2 used eye tracking as a prime modality of sensor input in their study. Therefore, on the \hlx{\texttt{Preparation of Study Materials}} card, they added that they would need a \hlpink{\textit{cover story specifically for explaining the use of eyetracking}} to potential members of the public.

The eyetracking also influenced \hlx{\texttt{Data Collection and Auxiliary Equipment}}. They personalized the card by adding the item \hlpink{\textit{synchronizing data inputs}} because it was necessary to ensure that \hlpink{\textit{each different sensing modality had synchronized time stamps}}. These are checklist item modifications specific to their study. 

After incorporating all of these changes, they can use the website to download their own personalized checklist and they can apply it throughout their study. P2 even said, ``I would have needed this checklist 2 years ago!''

\subsubsection{Checklist Evolution in Action}
\label{sec:thechecklist-inaction-evolution}

Our interviews also shed light on possible usages beyond individual research groups. For instance, the cards sparked discussion on the sequence of steps that the checklist presents that can be relevant for the wider community. 

The topic \hlx{Site Visits} in the set of topics  \textbf{\texttt{Before-Robot} } sparked  discussions that allowed us to refine an element of the checklist's high-level structure. In the version of the checklist that we presented to the experts, we had tried to order cards sequentially from left to right in each table cell. For some of the authors (A, D, E) and experts (P3), the robot influences the choice of site. In contrast, P4 shared that ``the site visit is the first thing that I would do.'' For this expert, the site influences the choice of robot. They further pointed out that providing a ``temporal sequence'' of the cards within the table cells, i.e. of topics beyond the general structure in before, during and after made the checklist less applicable to their research.  

The expert interviews helped us to further appreciate the flexibility that the card structure provides, and the discussion prompted us to update the checklist to reflect this. This experience demonstrated how a different researcher's perspective can further improve the checklist and spark discussion of practices in the community.%

\section{Discussion: Beyond the Checklist}
\label{sec:disc}

Our checklist is a powerful tool for the community and viewing it in conjunction with other facets of public HRI research produces valuable observations.

Despite the use of the checklist, unexpected and unplanned situations due to the nature of public deployments are unavoidable. For example, in Sydney's deployment, there were more people than expected at the site, which caused the robot to move more than anticipated. One of the robot's motors overheated because it had not been tested in a location with so many people. Several hours into the deployment, the motor needed time to cool down before the robot could be restarted. These kinds of minor (and sometimes major), site- and robot-specific problems occur in every deployment but don't always make it to a final report. 

Although challenges in robot deployments cannot be completely avoided, the checklist gives researchers the option to convert ``one-off'' experiences into a checklist modification to prepare their colleagues to not repeat their experiences. The ability to use topics to codify that tacit knowledge means that what was unexpected for this study can become expected and anticipated for future ones. In the folowing we discuss further implications.

\subsection{The Checklist and Design Principles}
As a community resource, the checklist can be a way to stimulate discussion about research paradigms and traditions.
For example, P4's participatory design goals for their deployment informed their prioritization. P4 established the site before the robot. During the participatory design process, site constraints influenced the choice of the robot and the design of its behaviors. 

In contrast, a researcher studying social navigation may first know what behaviors the robot will exhibit, then choose a site that allows them to study those behaviors. In both cases, researchers can use the checklist to consider the mutual influences that the robot and the site have on each other, but due to the varying objectives, the topics should be prioritized in different orders. 

Both cases represent valid applications of the checklist, but highlight how HRI in public is currently approached by different research groups. The checklist may help to articulate such differences. 

\subsection{Global Perspectives}
What is considered ethical or appropriate in one place, may be perceived as rude and imposing in another. Robot deployments tend to primarily be reported in western and industrialized countries~\cite{Winkle_Lagerstedt_Torre_Offenwanger_2022,Seaborn_Barbareschi_Chandra_2023} but even so, among our author team, split solely between North America and Europe, we were able to identify and discuss interesting cultural and legal differences. The geographic and cultural spread from our expert interviews incorporated even more perspectives that are different from our own, including perspectives from Asia and Australia allowing us to refine the checklist into an improved and more mature product. We have not yet had the opportunity to discuss the checklist with an expert with a global south perspective. We expect that as the involved group grows more diverse, the checklist will be refined into a form that is even more impactful. We would like for the checklist to be representative of a global community of public HRI researchers and seek to pursue this going into the future.

\subsection{Researcher Care} 
Some topics in the \texttt{Human} domain column that touch on the topic of researcher care, but it is far too complex of a concern to be wholly captured by several topics or bullet points. There have been recent calls for acknowledging the hidden labor in HRI that is expected from researchers \cite{Pelikanetal-wizards,Gamboaetal2025}. In two of our interviews, we specifically discussed the importance of preparing the researcher and robot handlers. There are practical issues related to the \textit{presentation of self}, including what clothes to wear to appear approachable or to turn invisible in the case of robot wranglers and wizards. It also includes elements of \textit{self-care}, taking sufficient breaks, staying safe, and being prepared for violence exercised by bystanders. But there can also be an \textit{affective dimension} that emerges when doing research in the wild, as building relationships with people outside the university may lead to unexpected emotions, including grief during and after the passing of study participants. We considered incorporating a new domain into the checklist-- a separate column specifically centered around the researchers or robot handlers but felt it would be reductive to extract it into an independent column. A concern that impacts the researcher's well-being will affect topics in the checklist in ways highly specific to each person's own experience. We chose to include topics in each phase of the \texttt{Human} domain to draw attention to researcher's health and safety and recommend checklist users to consider the gravity of these particular topics. We encourage members of the community to engage with literature on ethnography, primarily by anthropologists, which has covered concerns of doing research outside the lab and can provide practical guidelines related to safety and emotional preparation \cite{Procter_Spector_2024,Pollard-fieldwork,Bosco-fieldwork}. 

\section{Conclusion}
\label{sec:conclusion}
We developed a checklist for the deployment of public robots from the perspective of an interdisciplinary team of HRI researchers that work in a broad scope of areas of HRI (user studies, design, systems, technical and theory and methods). We convert the tacit knowledge of lab-specific pointers or personal experiences into a centralized resource. We demonstrated through a sample of interviews with field experts, a method to keep it alive and evolving. By providing an open-source, collaborative, flexible checklist, we take a step towards unifying the body of work and setting a community standard without being prescriptive. We hope this checklist begins a conversation in public HRI to help each other put robots in the real world.

\section{Acknowledgements}
The authors would like to thank Selma Šabanović, Zhe Zeng, Marius Hoggenmüller, Dražen Brščić, and Andrew Blair for their expertise they contributed to the checklist.

\balance

\bibliographystyle{ACM-Reference-Format}
\bibliography{BibFile}

\end{document}